\documentclass{bmvc2k}
\usepackage{subfigure}

\title{A Hybrid Supervised-unsupervised Method on Image Topic Visualization with Convolutional Neural Network and LDA}

\addauthor{Kai Zhen}{zhenk@indiana.edu}{1}
\addauthor{Mridul Birla}{mbirla@umail.iu.edu}{2}
\addauthor{David Crandall}{djcran@indiana.edu}{1}
\addauthor{Bingjing Zhang}{zhangbj@indiana.edu}{1}
\addauthor{Judy Qiu}{xqiu@indiana.edu}{1}

\addinstitution{
 School of Informatics and Computing,\\
 Indiana University\\
 Bloomington, Indiana
}
 \addinstitution{
  Yahoo Inc.\\
  1010 Yahoo Way,\\
  Seattle, Washington 
 }

\runninghead{Student, Prof, Collaborator}{Submission}

\def\etal{\emph{et al}\bmvaOneDot}

\begin{document}

\maketitle

\begin{abstract}
Given the progress in image recognition with recent data driven paradigms, it's still expensive to manually label a large training data to fit a convolutional neural network (CNN) model. This paper proposes a hybrid supervised-unsupervised method combining a pre-trained AlexNet with Latent Dirichlet Allocation (LDA) to extract image topics from both an unlabeled life-logging dataset and the COCO dataset. We generate the bag-of-words representations of an egocentric dataset from the softmax layer of AlexNet and use LDA to visualize the subject's living genre with duplicated images. We use a subset of COCO on $4$ categories as ground truth, and define consistent rate to quantitatively analyze the performance of the method, it achieves $84\%$ for consistent rate on average comparing to $18.75\%$ from a raw CNN model. The method is capable of detecting false labels and multi-labels from COCO dataset. For scalability test, parallelization experiments are conducted with Harp-LDA on a Intel Knights Landing cluster: to extract $1,000$ topic assignments for $241,035$ COCO images, it takes $10$ minutes with $60$ threads.
\end{abstract}

\section{Introduction}
\label{sec:intro}
With the prevalence of deep neural network in visual data nowadays, detecting patterns or recognizing objects from an image dataset has become less mysterious than it used to be. Given a manually labeled training data with sufficient images, we can tune the parameters of a convolutional neural network that yields near-perfect performance. However, those concepts are typically learned in a supervised setting. It's still time consuming and expensive to organize such a large volume of manual work to label the training data, especially that the amount of images generated by smart devices during a single day is significantly beyond the capacity of manual labor. We are wondering if those newly generated, unlabeled images can be classified, recognized, or labeled automatically, so that they can either be utilized for large-scale training purposes, or simply be better documented and organized in local devices.

\begin{figure}[h]
\centering
\includegraphics[width=0.97\textwidth]{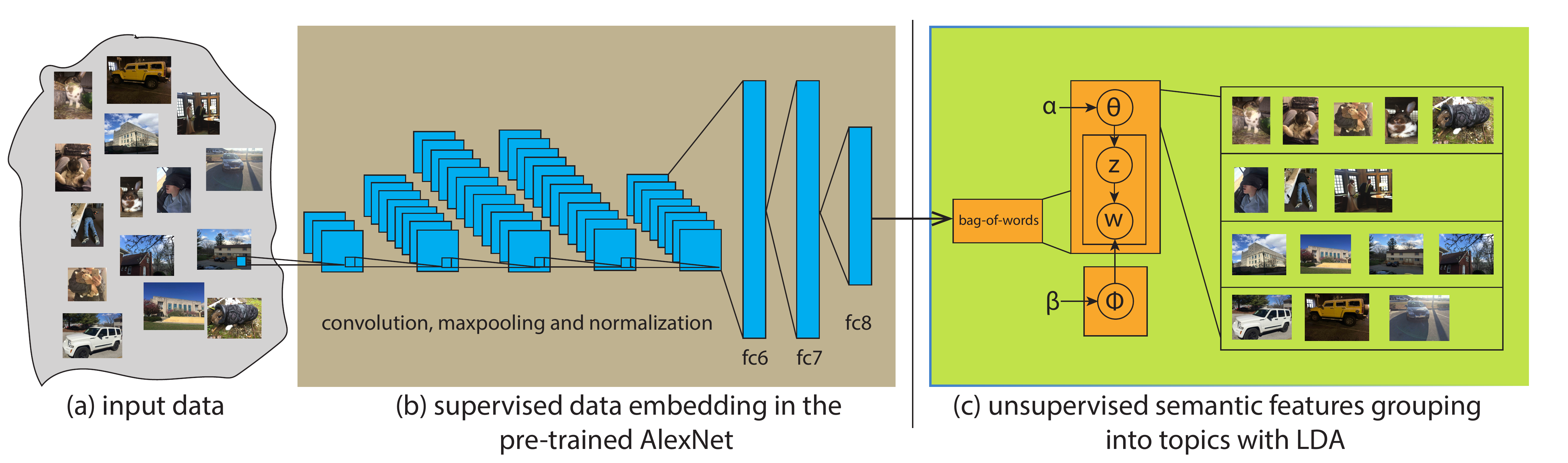}
\caption{Framework of the hybrid method to extract image topics: (a) after pre-processing the input data; (b) we put each image into a pre-trained AlexNet with ImageNet labels; (c) and then we run LDA on the generated bag-of-words representation to group low-leveled semantic features into visual topics.}
\label{fig:371}
\end{figure}

In this paper, we combine AlexNet and Latent Dirichlet Allocation (LDA) to form a hybrid supervised-unsupervised method to extract topics (visual concepts) from unlabeled datasets, see Fig.~\ref{fig:371}. The idea is to construct the embedding for each image from a universal pre-trained model, and then apply the topic model by grouping salient semantic features into visual topics. Concretely, we consider two scenarios. First, we take the challenge from a life-logging dataset. Life-logging cameras create huge collections of photos, 
even for a single person on a single 
day
~\cite{bambach2015lending,ryoo2013first,korayemenhancing}, which makes it difficult for users to browse or organize their
photos effectively. Unlike text corpora in which words create
intermediate representations that carry semantic meaning for
higher-level concepts such as topics, images have no such obvious
intermediate representation in between.
Egocentric
photos are particularly challenging because they were taken opportunistically,
so they are often blurry and poorly-composed compared to consumer-style
images. We use this method to ``summarize" a subject's living genre from an egocentric life-logging dataset.

Second, we use COCO dataset, a labeled dataset as ground truth to evaluate the hybrid method in terms of consistent rate. We apply LDA on top of the bag-of-word representation from a pre-trained AlexNet and get the topic assignment matrix over all images. By defining a concept of consistent rate, we compare the ground truth labels and ``concept clusters" from our method, and measure the consistency of those clusters. It shows that the space with $1,000$ ``irrelevant" dimensions suffice a dissimilarity measurement of image data by achieving an average consistent rate of $84\%$. We also apply Harp-LDA ~\cite{zhang2016high}, a parallel LDA based on sparse matrix decomposition, on the state of the art Intel Knights Landing cluster for parallization, which shows the feasibility for potential applications on scaled up experiment settings.


The method provides a hybrid way to extract image topics with a pre-trained AlexNet and a probabilistic topic model. It automatically labels images before manually double-checking, instead of having people label the entire image dataset; the living genre extracted from egocentric images can be used in potential psychological research; it can also detect duplicated images and organize photo albums in local computers. We describe the related work in section $2$, data in section $3$, methods in section $4$, experiment and results in section $5$, and summarize the work in section $6$.

\section{Related work}

Many methods have been proposed that adapt techniques from text topic modeling
to vision.
Li and Perona~\cite{fei2005bayesian} propose
a Bayesian hierarchical model to learn characteristic intermediate
themes in an unsupervised way, for example, while Sivic \etal~\cite{sivic2008unsupervised} and Li et al.~\cite{li2010building}
introduce ways to discover hierarchical image structure
from unlabeled
datasets~\cite{sivic2008unsupervised,li2010building}. 
However, 
most of these techniques were developed prior to the prevalence of deep neural network,
and are based on hand-tuned features. Moreover, none have studied
egocentric imagery, as we do here.

Research on computer vision for first-person images and video has been
popular in recent years, including in the fields of object
tracking~\cite{lee2014hand}, activity
recognition~\cite{pirsiavash2012detecting,iwashita2014first}, and
event detection~\cite{lu2013story}. However, there has
been relatively little work on unsupervised object discovery and scene
summarization in this domain~\cite{ryoo2013first}. Here we
take a first step towards understanding the extent to which existing
hierarchical Bayesian topic models paired with deeply-learned features
could be successfully applied to the unique properties of first-person
imagery, such as repetitive scenes, frequent motion blur, and poor
image composition.

\section{Data}
\label{section:res}
We are interested in extracting topics from egocentric images as a way to visualize the subject's living genres. This is assumed to be relatively feasible as it may contain large amount of duplicated images (for example, a series of images on a subject sitting in front of a laptop are with sufficient similarity to be considered duplicated). Then we test  the method on COCO dataset with quantitative analysis on accuracy and scalability.
\subsection{Lifelogging dataset.}
The dataset of first-person images is captured with a Narrative Clip lifelogging camera by one of the authors.
We wore that camera, which
takes pictures about every $30$ seconds, for two weeks during
Summer $2015$. The camera captured $7,927$ ($12$ days) images
of a wide variety daily activities including commuting
to work, having meetings, preparing and eating meals, interacting
with friends and family, etc. The lifelogging user reviewed
all of the images after collection and removed about
$20$ that they felt too private to share.

\subsection{COCO dataset}
In order to evaluate the performance quantitatively and  test the scalability of the hybrid supervised-unsupervised method, we use Microsoft COCO (Common Object in 
Context) dataset~\cite{lin2014microsoft}, as it contains $80$ categories in $12$ super-categories with over $200,000$ images in total (some images are associated with multiple categories). We randomly select a subset of COCO with $4$ categories, compute and compare consistent rate with other methods. The whole dataset is used for scalability experiment.

\section{Methods}
\label{section:met}
As with previous works, we assume that features from the layer of CNN can be regarded as visual words which compose images in a similar way as words compose documents in text corpora, and that the ordering factor of visual words can be ignored. Empirical studies show that the occurrence of words in each documents can be modeled as multinomial distribution~\cite{nigam2000text,mccallum1998comparison}. 

\subsection{Data preprocessing and pre-trained model}
To get the order-irrelevant (the order of ``words" itself in each image carries no information) representation and learn a joint distribution of images and topics, we process the data in three steps shown in Fig.~\ref{fig:371}. We first use a pre-trained AlexNet (trained on ImageNet) to extract the label response of each image from the output layer. We choose AlexNet, instead of more recent architectures, as the pre-trained model as it is pervasively used in current research and applications.
The representation is extracted from the softmax layer, which is an $1,000$ dimensional vector corresponding to $1,000$ labels from AlexNet~\cite{krizhevsky2012imagenet}. Each probability is with the range of $[-10, 10]$ that shows how likely the image is related to the corresponding label. The greater probability indicates a larger likelihood. 

We set a threshold and keep the indices those labels with probabilities above the threshold. This removes labels that are less semantically salient and gives a ``bag-of-words" representation where the data order is irrelevant. After preprocessing, each image is represented with a vector of label IDs that are bounded between $0$ and $999$. With this representation, we build a classic LDA model to learn a distribution over topics for each image.

\subsection{Generative model and collapsed Gibbs sampler}

We model the relation between a feature to an image as a word to a document, and assume that there is a hidden variable of topic in between words and images. In a similar generative process as proposed in Latent Dirichlet Allocation~\cite{blei2003latent}, an image is generated by first assigning topics and then sampling features (visual words) from selected topics which is given in Eq.~\ref{eq2}. There are three parameters, i.e. $z$, $\theta$, $\phi$ to be inferred from the posterior distribution, where $\alpha$ and $\beta$ are hyperpriors, $\theta$ is the distribution over topics for each document (image), $\phi$ is the distribution over words for each topic, and $z$ is the topic allocations.

\begin{equation} \label{eq2}
\begin{split}
p(w,z,\theta,\phi|\alpha,\beta)=p(\theta|\alpha)p(z|\theta)p(\phi|\beta)p(w|\phi,z)
\end{split}
\end{equation}

Collapsed Gibbs sampling works as $\phi$ and $\theta$ can be represented in a closed form by $z$, see Eq.~\ref{eq3}~\cite{griffiths2004finding}. Here $n(i,z)$ is the count of words in images $i$ being assigned to topic $z$; $n(z,w)$ means the total number of the word $w$ being assigned to the topic $z$; $Z$ and $W$ are the total number of topics and distinct words, respectively. In this work, we are particularly interested in $\Theta$ matrix through which we visualize the topic $z$ by displaying representative images (in practice, we pick top $5$ images).
\begin{equation} \label{eq3}
\begin{split}
\theta_{i,z} &= \frac{n(i,z)+\alpha}{\sum_Z (n(i,z)+\alpha)}\\
\phi_{z,w} &= \frac{n(z,w)+\beta}{\sum_W (n(z,w)+\beta)}
\end{split}
\end{equation}

The collapsed Gibbs sampler needs to calculate the probability of the $n^{th}$ word in the $m^{th}$ image being assigned to a topic $k$, given all other topic assignments to the rest words in all images. Integrate out multinomial parameters and we calculate that probability based on Eq.~\ref{eq4}. In Eq.~\ref{eq4}, $-(m,n)$ means all words but the $n^{th}$ word in the $m^{th}$ image. For example, $C_{k,m,*}^{-(m,n)}$ means the count of all the words in image $m$ that have been assigned to topic $k$, except the $n^{th}$ word; $C_{k,*,*}^{-(m,n)}$ means the count for all words from all images with the topic assignment of $k$, except the $n^{th}$ word in the $m^{th}$ image. In practice, the term $C_{k,*,*}^{-(m,n)}$ is dropped out as it is a constant in each image.

\begin{equation} \label{eq4}
\begin{split}
p(z_{(m,n)} = k|z_{-(m,n)},w,\alpha,\beta) \approx \\
 \frac{\alpha+C_{k,m,*}^{-(m,n)}}{Z*\alpha+C_{*,m,*}^{-(m,n)}}
 *\frac{\beta+C_{k,*,w_{m,n}}^{-(m,n)}}{W*\beta+C_{k,*,*}^{-(m,n)}}\approx \\
  \frac{(\alpha+C_{k,m,*}^{-(m,n)})(\beta+C_{k,*,w_{m,n}}^{-(m,n)})}{W*\beta+C_{k,*,*}^{-(m,n)}}
\end{split}
\end{equation}

\section{Experiment and results}
We conducted the ``bag-of-words" representation part on Dell PowerEdge $T630$ server with two NVidia Tesla $K40$ GPU boards via Caffe~\cite{jia2014caffe}. For lifelogging dataset, it took approximately $200$ minutes to generate the corpus. For COCO dataset, it took $40$ hours in practice. We implemented sequential version of LDA for lifelogging data and a subset of COCO data for accuracy evaluation. To test the scalability, we ran Harp-LDA~\cite{zhang2016high} on on Juliet, a cluster for digital science computing as a part of FutureSystems in Indiana University. In practice we conducted Harp-LDA experiment on a cluster of $2$ machines with $24$ cores each. 
\subsection{Lifelogging data summarization with duplicated images}

\begin{figure}[h!]
\centering
\subfigure[$2$ topics with $5$ image samples.]{\includegraphics[width = 2in]{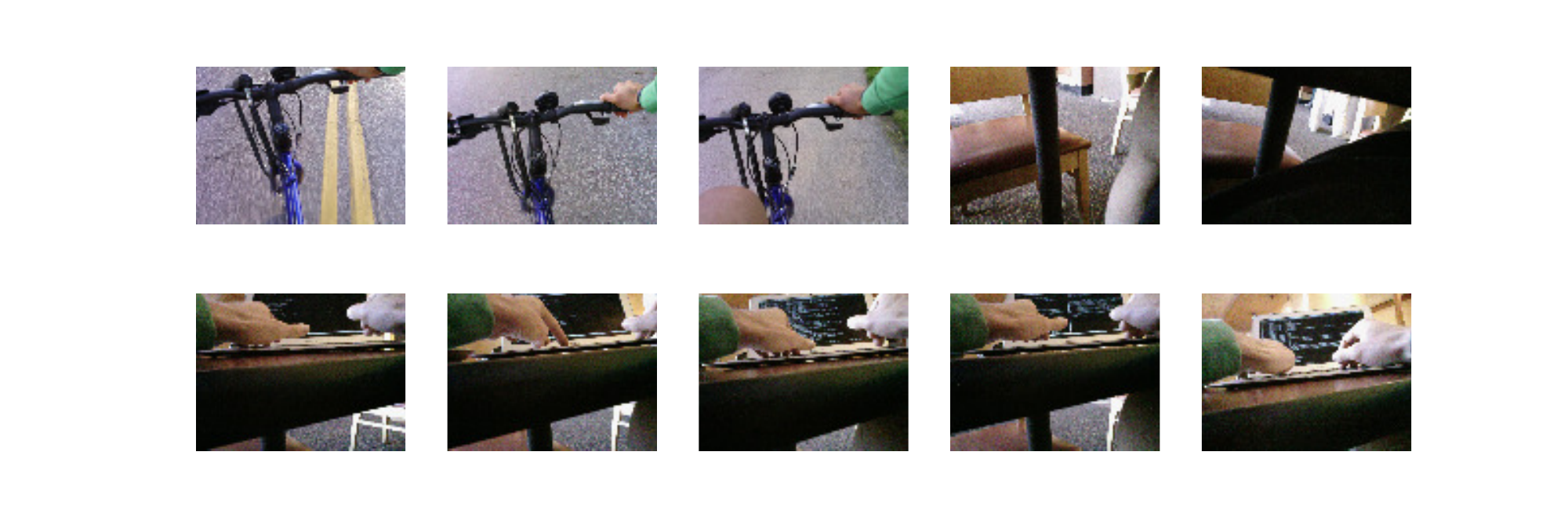}}
\subfigure[$3$ topics with $5$ image samples.]{\includegraphics[width = 2in]{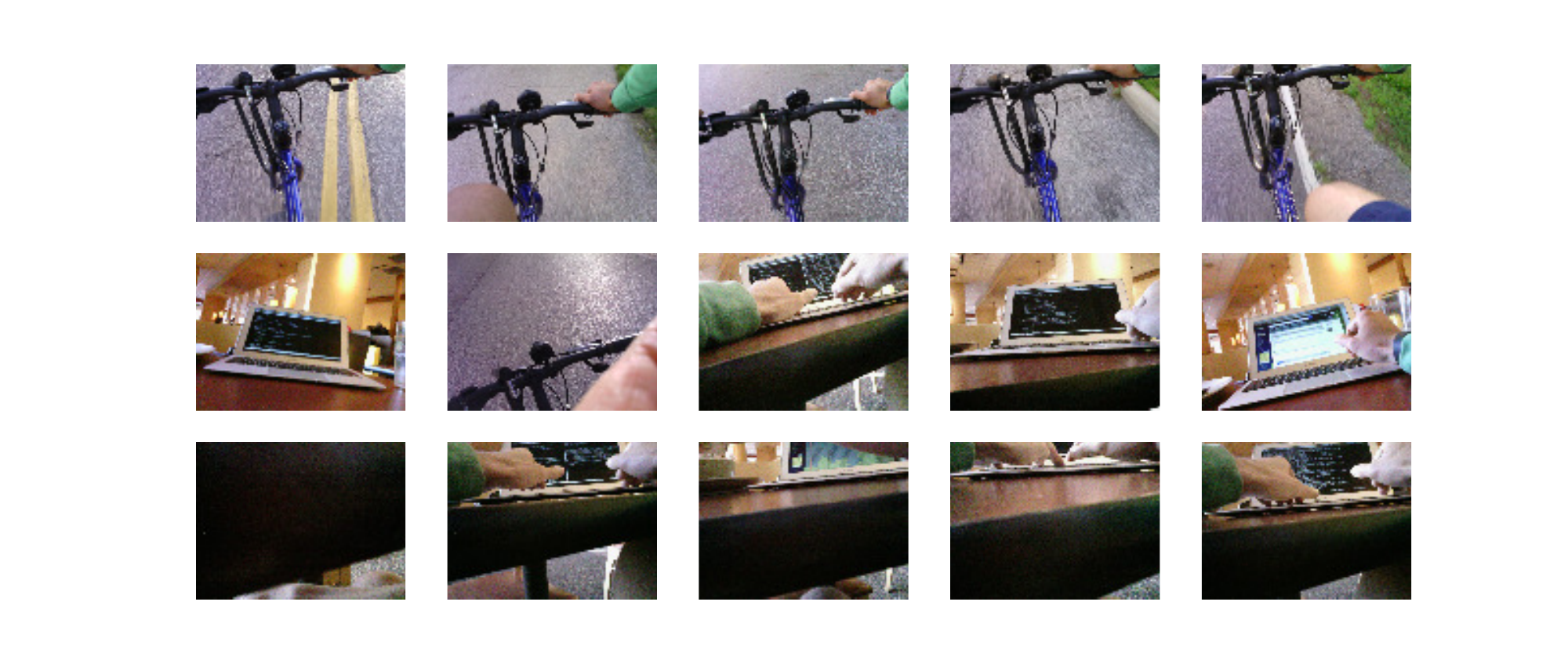}}
\caption{Topic visualization of egocentric images on a single day (when choosing the number of topics that is greater than the actual daily activities, say $5$ topics on a workday with coding and riding bike, these activities will occur accross more than one topic).}
\label{fig:223}
\end{figure}

The method was applied on the egocentric dataset of a single school day where the subject ride to school and work. We extracted $2$ topics and $3$ topics respectively from the dataset and select the top $5$ images from each topic, as shown in Fig.~\ref{fig:223}.

In the case of $2$-topics, the T-shaped bicycle is bonded with T-shaped tables and the second topic indicates a working scenario with the laptop. Given $3$ topics, the grouping of semantic features is conducted in a finer granularity, as it decomposes the second topic from $2$-topic-case into a topic related to the computer screen and another topic for the desk. The first topic has more purity with all top $5$ images being bicycles. 

\begin{figure}[h!]
\centering
\includegraphics[width=0.8\textwidth]{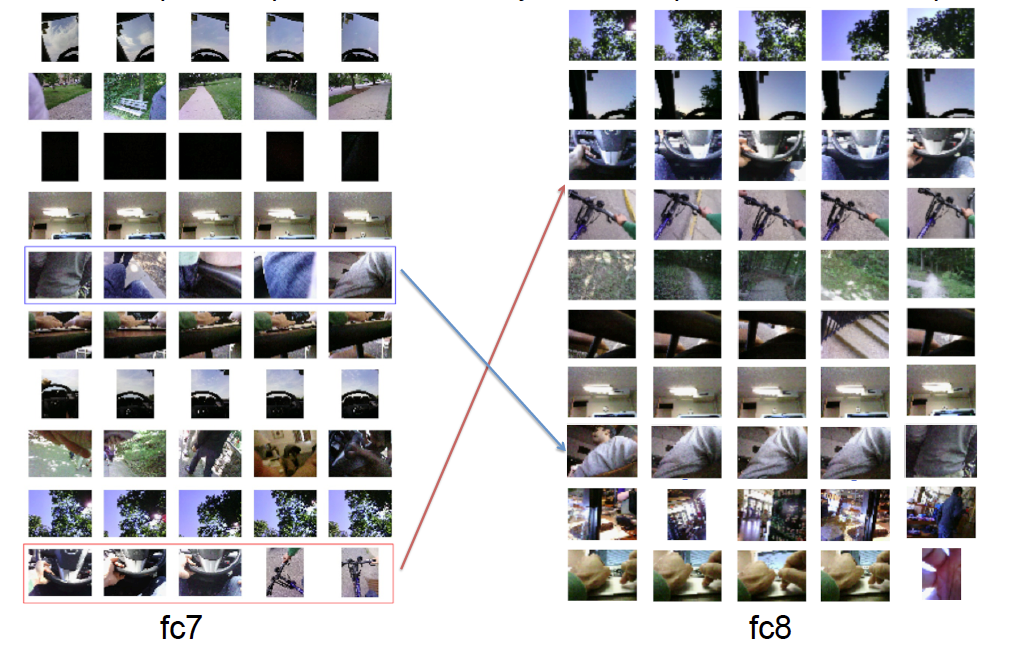}
\caption{$10$ topics extracted from egocentric images on a week. The ``bag-of-words" representation is constructed from both $fc7$ (left) and $fc8$ (right) layers and results are displayed in parallel.}
\label{fig:444}
\end{figure}

We generalized the method to a larger dataset for weekly images and selected $10$ topics with $5$ top representative images. The results based on both $fc7$ and $fc8$ layers are shown in Fig.~\ref{fig:444}.  Both representations yield a similar living genrethe with topics on driving, meeting with collegues, working with the laptop, walking in the yard, etc. The output from softmax layer leads to marginally more coherent semantic groupings.
%

\begin{figure}[h!]
\centering
\includegraphics[width=0.75\textwidth]{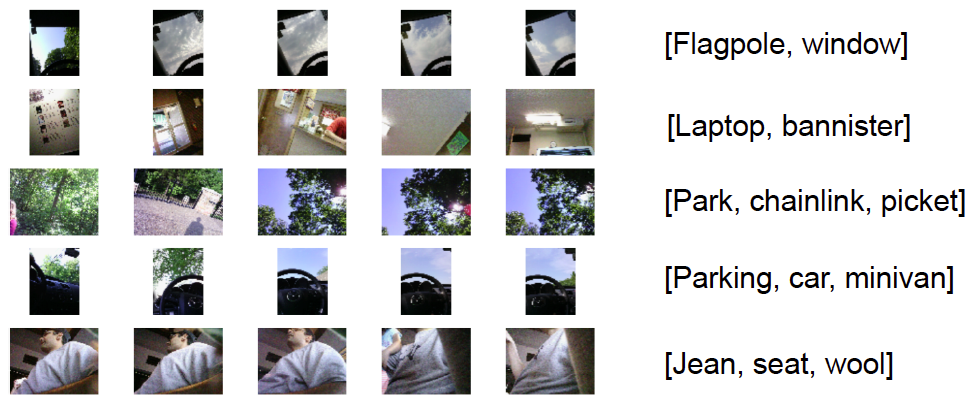}
\caption{Topics in life-logging data and corresponding labels from pre-trained AlexNet (the model is trained on ImageNet with $1,000$ labels that are not specifically for the egocentric dataset).}
\label{fig:44}
\end{figure}

While the hybrid supervised-unsupervised method generates a set of topics each of which conveys consistent semantic meaning, the original labels, derived from ImageNet, are not highly corelated to the content of our life-logging dataset.
As shown in Fig.~\ref{fig:44} where we displayed corresponding labels of each topic from AlexNet, these labels may not form appropriate captions for the corresponding images in each topic. However, the high dimensional space constructed with $1,000$ labels suffices the embedding of the semantic dissimilarity. In the next section, we calculate consistent rate to evaluate the method with a subset of COCO data with ground truth labels.

\subsection{Topic spectrogram and consistent rate}
\begin{figure}[h!]
\centering
\includegraphics[width=0.9\textwidth]{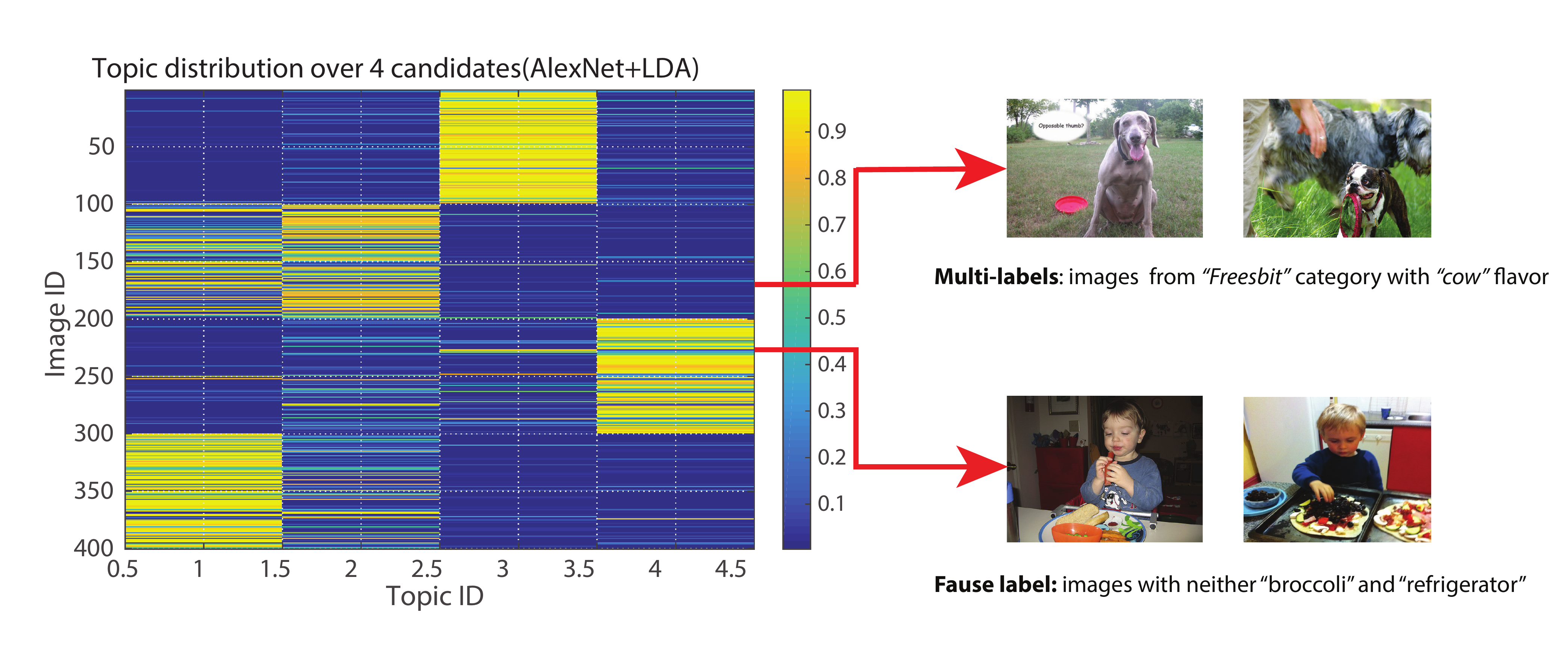}
\caption{Spectrogram of $4$ topics learned from the subset of COCO: we can see a clear grid layout in this spectrogram. For the first and third topics, most of the images are assigned to one topic with a predominant probability. For the $2nd$ and $4th$ topics on ``Frisbee" and ``Cow", they are mutually interrelated. }
\label{spec1}
\end{figure}

To have a quantitative analysis on the performance of this method, we first randomly selected $400$ images among $4$ categories (broccoli, frisbee, fridge and cow) from COCO, $100$ images for each category, extracted the bag-of words representation from CNN, and calculated the topic distribution by LDA, as is shown in Fig.~\ref{spec1}. Overall, we see a clear topic assignment ``mode" for every $100$ images. Note that the label set of ImageNet does not include ``Frisbee" and ``Cow". With the related labels, such as ``French\_bulldog", ``Bernese\_mountain\_dog", ``lawn\_mower" etc, the method built the synthesized semantic labels that could group images in those two categories. Some images from those two categories are correlated semantically: dogs are likely to be seen with a frisbee in a scenario of outdoor activities. From the spectrogram, we see that images responding to both topics contain both dogs and frisbee. For the $1st$ and $3rd$ topic where topic assignment is more consistent, if a topic assignment of one image is different from the majority in that category, it's likely that the image was not labeled correctly (we listed two images that should have contained either brocolli and a refrigerator).

\begin{figure}[h!]
\centering
\includegraphics[width=0.97\textwidth]{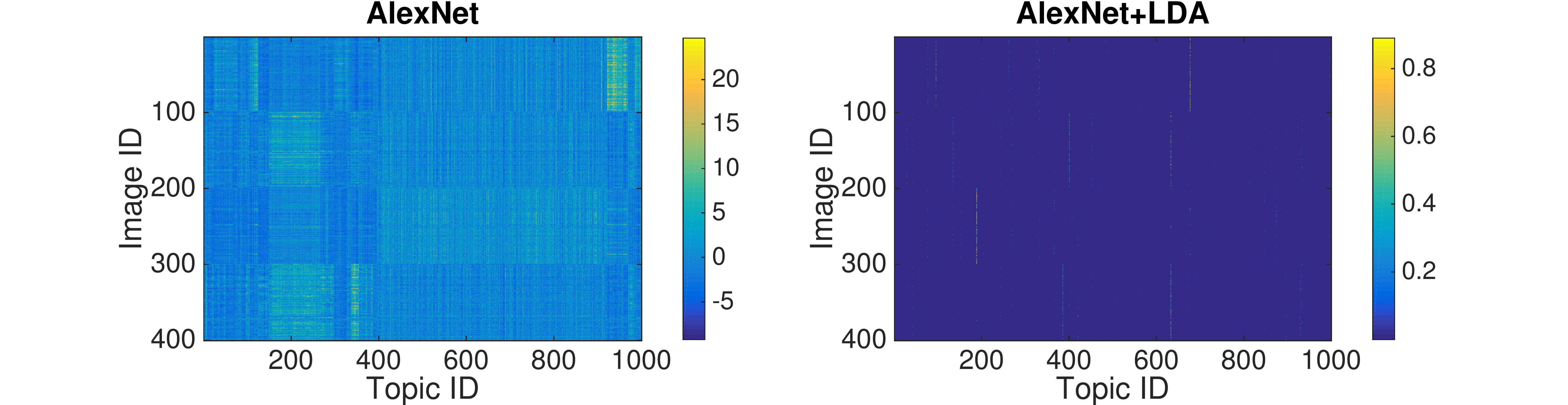}
\caption{Topic distribution comparison between AlexNet and AlexNet+LDA: since the AlexNet is pre-trained on ImageNet with a set of labels different from those in COCO image dataset, for each image from COCO, there are many responding labels. However, LDA helps to soak up the noise and leave only a few responding labels, even it's also running on $1,000$ topics.}
\label{spec}
\end{figure}

We compared the output from pure CNN and CNN+LDA by extending the number of topic of LDA to $1,000$, the same amount of labels from the output layer of AlexNet trained on ImageNet.
 The spectrogram of the output from the softmax layer of AlexNet, and compared that with the output of LDA on the same amount of topics is shown in Fig.~\ref{spec}.


To measure the consistent rate of the output labels, we calculated the index with the largest mean probability among $1,000$ labels for each image subset $(\bar{p}_{max})$, and counted the number of images the label with the largest probability equal to the corresponding index $I(p(i)=\bar{p}_{max})$, and divided it by the amount of images in the subset, i.e. $I(p(i)=\bar{p}_{max})/\#images$. The result is shown in Table ~\ref{my-label}. 

\begin{table}[h!]
\centering
\caption{consistent rate: the number $i$ in the bracket means the top $i$ labels considered. For CNN+LDA[1], only the label with the largest probability is calculated.}
\label{my-label}
\begin{tabular}{||l||l|l|l|l||}
 \hline
 Methods & Broccoli& Frisbee& Fridge &Cow\\
 \hline
CNN         &0.3500 &   0.1100 &   0.1500 &   0.1400 \\
CNN+LDA[1] &0.6300 &   0.4400 &   0.7500 &   0.4900 \\
CNN+LDA[2] &0.8600 &   0.7500 &   0.8000 &   0.7600 \\
CNN+LDA[3] &0.8900 &   0.8400 &   0.8600 &   0.7700 \\
\hline
\end{tabular}
\end{table}

The consistent rate of CNN is relatively low with a wide range of responding labels. With LDA to extract topics from labels, the consistent rate is improved by $3.08$ times on average. For the case of CNN+LDA[1], the consistent rate of ``frisbee" and ``fridge" is much lower comparing to the other categories because those two categories are correlated: some images under ``frisbee" was assigned to ``fridge" and vice versa. When $k>1$ for top $k$ largest probabilities, the consistent rate for the two categories became close to those of other two categories.

\subsection{Parallelization for scaled-up dataset}
\begin{figure}[h!]
\centering
\includegraphics[width=0.5\textwidth]{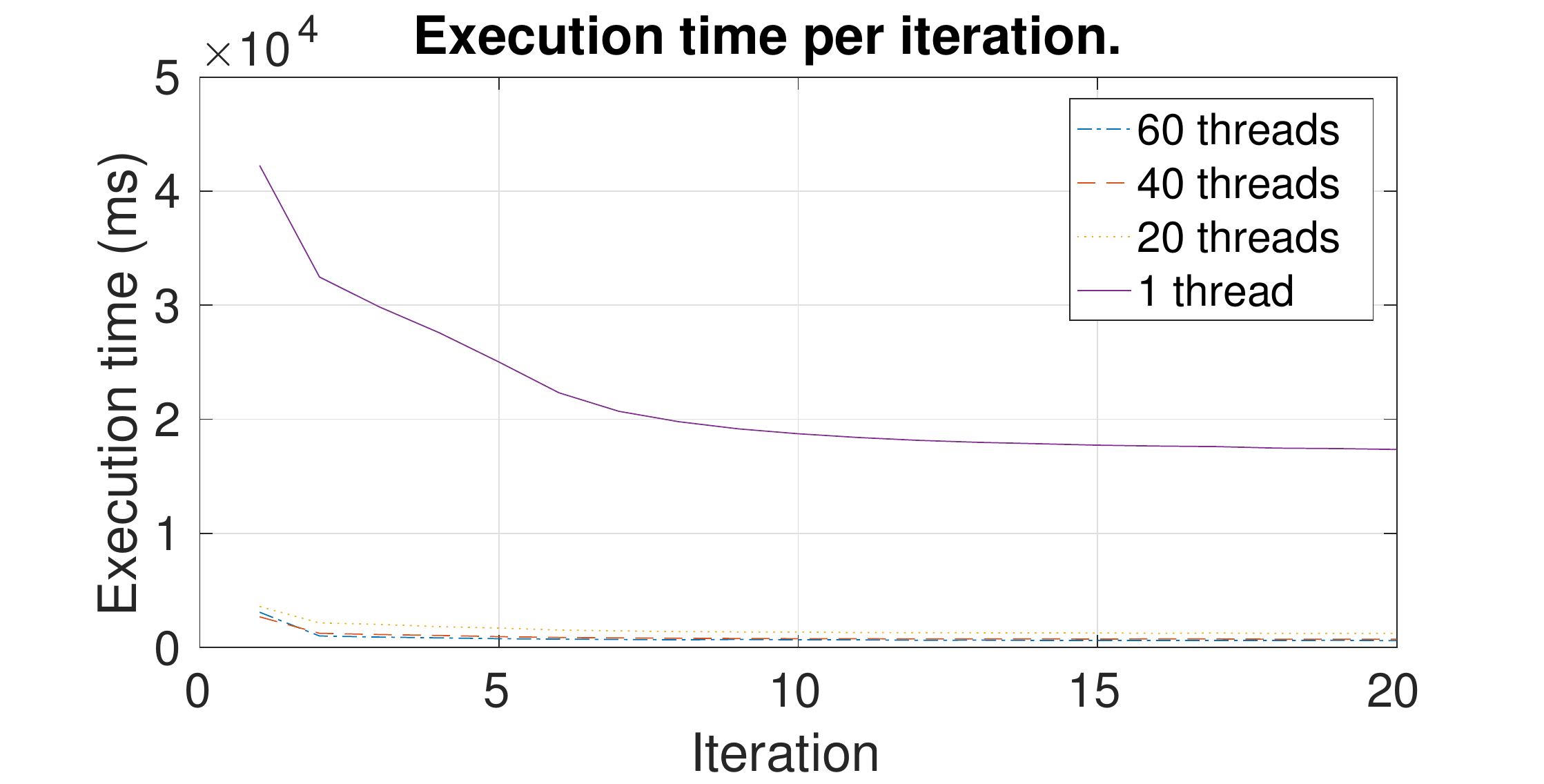}
\caption{Execution time for the first $20$ iterations of Gibbs sampling algorithm on KNL machine given different numbers of threads.}
\label{spec314}
\end{figure}

Running a preprocessed COCO image dataset with $1.8M$ tokens from $241,035$ images for $1,000$ topics is beyond the computing capacity of most local computers. Instead, we ran the experiment on a Intel Knights Landing cluster at Indiana University.
One node with Xeon Phi $7250$F processors ($68$ cores in total) was used.
All of them have around $200$ GB memory and are connected by Omni-Path Fabric. We used Harp-LDA ~\cite{zhang2016high} based on sparse matrix decomposition, recorded the execution time for each iterations and estimated the overall execution time for $1,000$ iterations, see Fig.~\ref{spec314}. 


The execution time for each experiment setting will converge after $20$ iterations. When using $1$ thread in a single node, the overall execution time for $1,000$ is $16,139,258$ ms (around $4$ hours and $30$ minutes). With $20$ threads, it's $1,235,859$ ms; for $40$ and $60$ threads, they are $814,546$ ms and $625,012$ ms (around $10$ minutes) respectively.

\section{Conclusion}
The paper showed that the image feature generated by Convolutional Neural Network contains semantic information that can be grouped and extracted by topic modeling in a supervised-unsupervised way. The pre-trained network with a ``non-specific" set of labels gave the bag-of-words representation which measures dissimilarity yet with low purity. With topic modeling, we further grouped the semantic features into visual topics of higher level. The method was applied on our life-logging dataset to extract living genres by listing duplicated images, and on a subset of COCO dataset for consistent rate analysis. We also conducted a parallel experiment on KNL machine with all COCO images over $80$ categories for scalability test. The topic assignment procedure for $241,035$ images costs $10$ minutes. The method can be used to extract living genres from egocentric data with duplicated images; automatically group, pre-label and semantically organize images by topics.

\bibliography{zhenkbmvc}

\begin{thebibliography}{18}
\providecommand{\natexlab}[1]{#1}
\providecommand{\url}[1]{\texttt{#1}}
\expandafter\ifx\csname urlstyle\endcsname\relax
  \providecommand{\doi}[1]{doi: #1}\else
  \providecommand{\doi}{doi: \begingroup \urlstyle{rm}\Url}\fi

\bibitem[Bambach et~al.(2015)Bambach, Lee, Crandall, and
  Yu]{bambach2015lending}
Sven Bambach, Stefan Lee, David~J Crandall, and Chen Yu.
\newblock Lending a hand: Detecting hands and recognizing activities in complex
  egocentric interactions.
\newblock In \emph{Proceedings of the IEEE International Conference on Computer
  Vision}, pages 1949--1957, 2015.

\bibitem[Blei et~al.(2003)Blei, Ng, and Jordan]{blei2003latent}
David~M Blei, Andrew~Y Ng, and Michael~I Jordan.
\newblock Latent dirichlet allocation.
\newblock \emph{the Journal of Machine Learning Research}, 3:\penalty0
  993--1022, 2003.

\bibitem[Fei-Fei and Perona(2005)]{fei2005bayesian}
Li~Fei-Fei and Pietro Perona.
\newblock A bayesian hierarchical model for learning natural scene categories.
\newblock In \emph{Proceedings of the IEEE Conference on Computer Vision and
  Pattern Recognition}, volume~2, pages 524--531, 2005.

\bibitem[Griffiths and Steyvers(2004)]{griffiths2004finding}
Thomas~L Griffiths and Mark Steyvers.
\newblock Finding scientific topics.
\newblock \emph{Proceedings of the National Academy of Sciences}, 101\penalty0
  (suppl 1):\penalty0 5228--5235, 2004.

\bibitem[Iwashita et~al.(2014)Iwashita, Takamine, Kurazume, and
  Ryoo]{iwashita2014first}
Yumi Iwashita, Asamichi Takamine, Ryo Kurazume, and MS~Ryoo.
\newblock First-person animal activity recognition from egocentric videos.
\newblock In \emph{2014 22nd International Conference on Pattern Recognition
  (ICPR)}, pages 4310--4315. IEEE, 2014.

\bibitem[Jia et~al.(2014)Jia, Shelhamer, Donahue, Karayev, Long, Girshick,
  Guadarrama, and Darrell]{jia2014caffe}
Yangqing Jia, Evan Shelhamer, Jeff Donahue, Sergey Karayev, Jonathan Long, Ross
  Girshick, Sergio Guadarrama, and Trevor Darrell.
\newblock Caffe: Convolutional architecture for fast feature embedding.
\newblock In \emph{Proceedings of the ACM International Conference on
  Multimedia}, pages 675--678. ACM, 2014.

\bibitem[Korayem et~al.()Korayem, Templeman, Chen, and
  Kapadia]{korayemenhancing}
Mohammed Korayem, Robert Templeman, Dennis Chen, and David Crandall~Apu
  Kapadia.
\newblock Enhancing lifelogging privacy by detecting screens.

\bibitem[Krizhevsky et~al.(2012)Krizhevsky, Sutskever, and
  Hinton]{krizhevsky2012imagenet}
Alex Krizhevsky, Ilya Sutskever, and Geoffrey~E Hinton.
\newblock Imagenet classification with deep convolutional neural networks.
\newblock In \emph{Advances in neural information processing systems}, pages
  1097--1105, 2012.

\bibitem[Lee et~al.(2014)Lee, Bambach, Crandall, Franchak, and Yu]{lee2014hand}
Stefan Lee, Sven Bambach, David Crandall, John Franchak, and Chen Yu.
\newblock This hand is my hand: A probabilistic approach to hand disambiguation
  in egocentric video.
\newblock In \emph{Proceedings of the IEEE Conference on Computer Vision and
  Pattern Recognition Workshops}, pages 543--550, 2014.

\bibitem[Li et~al.(2010)Li, Wang, Lim, Blei, and Fei-Fei]{li2010building}
Li-Jia Li, Chong Wang, Yongwhan Lim, David~M Blei, and Li~Fei-Fei.
\newblock Building and using a semantivisual image hierarchy.
\newblock In \emph{Proceedings of the IEEE Conference on Computer Vision and
  Pattern Recognition}, pages 3336--3343, 2010.

\bibitem[Lin et~al.(2014)Lin, Maire, Belongie, Hays, Perona, Ramanan,
  Doll{\'a}r, and Zitnick]{lin2014microsoft}
Tsung-Yi Lin, Michael Maire, Serge Belongie, James Hays, Pietro Perona, Deva
  Ramanan, Piotr Doll{\'a}r, and C~Lawrence Zitnick.
\newblock Microsoft coco: Common objects in context.
\newblock In \emph{European Conference on Computer Vision}, pages 740--755.
  Springer, 2014.

\bibitem[Lu and Grauman(2013)]{lu2013story}
Zheng Lu and Kristen Grauman.
\newblock Story-driven summarization for egocentric video.
\newblock In \emph{Proceedings of the IEEE Conference on Computer Vision and
  Pattern Recognition}, pages 2714--2721, 2013.

\bibitem[McCallum et~al.(1998)McCallum, Nigam, et~al.]{mccallum1998comparison}
Andrew McCallum, Kamal Nigam, et~al.
\newblock A comparison of event models for naive bayes text classification.
\newblock In \emph{AAAI-98 workshop on learning for text categorization},
  volume 752, pages 41--48. Citeseer, 1998.

\bibitem[Nigam et~al.(2000)Nigam, McCallum, Thrun, and Mitchell]{nigam2000text}
Kamal Nigam, Andrew~Kachites McCallum, Sebastian Thrun, and Tom Mitchell.
\newblock Text classification from labeled and unlabeled documents using em.
\newblock \emph{Machine learning}, 39\penalty0 (2-3):\penalty0 103--134, 2000.

\bibitem[Pirsiavash and Ramanan(2012)]{pirsiavash2012detecting}
Hamed Pirsiavash and Deva Ramanan.
\newblock Detecting activities of daily living in first-person camera views.
\newblock In \emph{Computer Vision and Pattern Recognition (CVPR), 2012 IEEE
  Conference on}, pages 2847--2854. IEEE, 2012.

\bibitem[Ryoo and Matthies(2013)]{ryoo2013first}
Michael Ryoo and Larry Matthies.
\newblock First-person activity recognition: What are they doing to me?
\newblock In \emph{Proceedings of the IEEE Conference on Computer Vision and
  Pattern Recognition}, pages 2730--2737, 2013.

\bibitem[Sivic et~al.(2008)Sivic, Russell, Zisserman, Freeman, and
  Efros]{sivic2008unsupervised}
Josef Sivic, Bryan~C Russell, Andrew Zisserman, William~T Freeman, and Alexei~A
  Efros.
\newblock Unsupervised discovery of visual object class hierarchies.
\newblock In \emph{Proceedings of the IEEE Conference on Computer Vision and
  Pattern Recognition}, pages 1--8, 2008.

\bibitem[Zhang et~al.(2016)Zhang, Peng, and Qiu]{zhang2016high}
Bingjing Zhang, Bo~Peng, and Judy Qiu.
\newblock High performance lda through collective model communication
  optimization.
\newblock \emph{Procedia Computer Science}, 80:\penalty0 86--97, 2016.

\end{thebibliography}
\end{document}